\documentclass{article}
\usepackage{spconf,amsmath,graphicx}
\usepackage{xcolor}
\usepackage{amsfonts} 
\usepackage{makecell, rotating, multirow, diagbox}
\usepackage{arydshln}
\usepackage{url}
\usepackage{hyperref}

\def\ie{{\em i.e.}}
\begin{document}
\title{Fusing Global and Local Features for Generalized AI-synthesized Image Detection}
%
\name{Yan Ju$^\dagger$, Shan Jia$^\dagger$, Lipeng Ke$^\dagger$, Hongfei Xue$^\dagger$, Koki Nagano$^\ddagger$, and Siwei Lyu$^\dagger$}
\address{
$^\dagger$University at Buffalo, State University of New York\\
$^\ddagger$NVIDIA}
%
%

%
%
\maketitle
\begin{abstract}
With the development of the Generative Adversarial Networks (GANs) and DeepFakes, AI-synthesized images are now of such high quality that humans can hardly distinguish them from real images. It is imperative for media forensics to develop detectors to expose them accurately. Existing detection methods have shown high performance in generated images detection, but they tend to generalize poorly in the real-world scenarios, where the synthetic images are usually generated with unseen models using unknown source data. In this work, we emphasize the importance of combining information from the whole image and informative patches in improving the generalization ability of AI-synthesized image detection. Specifically, we design a two-branch model to combine global spatial information from the whole image and local informative features from multiple patches selected by a novel patch selection module. Multi-head attention mechanism is further utilized to fuse the global and local features. We collect a highly diverse dataset synthesized by $19$ models with various objects and resolutions to evaluate our model. Experimental results demonstrate the high accuracy and good generalization ability of our method in detecting generated images. Our code is available at \url{https://github.com/littlejuyan/FusingGlobalandLocal}. 

\end{abstract}
\vspace{-0.2cm}

\vspace{-0.45cm}
\section{Introduction}
\label{sec:intro}
\vspace{-0.35cm}

Recently, the creation of realistic synthetic media has seen rapid developments with ever improving visual qualities and run time efficiencies. Notable examples of AI-synthesized media include the high-quality images synthesized by the Generative Adversarial Networks~(GANs)~\cite{karras2020analyzing}, and videos with face swaps or puppetry created by the auto-encoder models~\cite{li2020celeb}.
The synthetic images/videos have become challenging for humans to distinguish~\cite{guo2022open}, and correspondingly, a slew of detection methods~\cite{wang2022gan,lyu2020deepfake} have been developed to mitigate the potential risks posed by such AI-synthesized images. 


\begin{figure}[!t] 
\centering 
\includegraphics[width=0.48\textwidth]{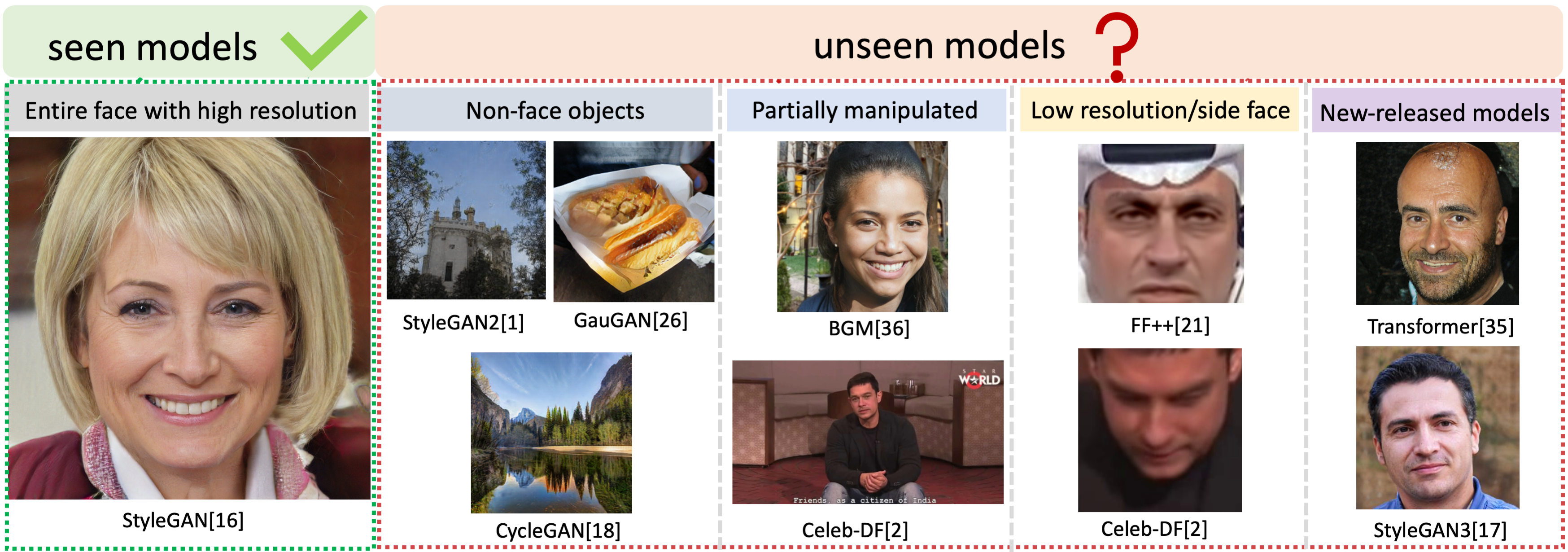}
\vspace{-0.8cm}
\caption{\small \em Detectors trained with high resolution frontal face images generated by the seen model have high accuracies. What about the images with various resolutions, objects and manipulation types from unseen models?} 
\vspace{-0.7cm}
\label{fig:examples} 
\end{figure}

Existing detection methods generally employ deep neural networks to automatically extract discriminative features, and then feed them into a binary classifier to tell real or fake. Most of the methods focus on global features extracted from the entire images~\cite{wang2020cnn,zhao2021multi,gragnaniello2021gan,asnani2021reverse}. Inspired by the success in fine-grained classification field, a recent work~\cite{zhao2021multi} brings a novel perspective by taking the binary image detection as a fine-grained classification task, and proposes to capture local features from multiple face attentive regions.

Current detectors have provided an excellent performance in a closed~(\ie{}, intra-dataset) scenario, where the testing and training images are generated by the same models trained on similar source data. However, their performance tends to decrease considerably in the open-world applications, where the testing images come from a different domain, and generated by unknown models or from unseen source data. For example, \cite{hulzebosch2020detecting} evaluates several SOTA detectors under the real-world conditions, including cross-model, cross-data, and post-processing, and concludes that current detection models are not yet robust enough for real applications.

To improve the detection generalization, several recent works propose to utilize new deep learning techniques, such as transfer learning~\cite{jeon2020t}, incremental learning~\cite{marra2019incremental}, contrastive learning~\cite{cozzolino2021towards}, representation learning~\cite{kim2021fretal}, etc. Data augmentation is also a successful tool to improve the generalization ability and robustness \cite{wang2020cnn}. Wang et al.~\cite{wang2020cnn} demonstrate that a standard classifier can generalize well if trained with careful pre-/post-processing and data augmentation.  

However, as the manipulation methods are rapidly developed and improved, how the existing detectors will generalize to various AI-synthesized images (as shown in Fig.~\ref{fig:examples}) deserves deep investigation. On the one hand, advanced generation models have made the differences between real and fake images more subtle and local. On the other hand, different types of manipulation with either global synthesis or local manipulation are increasing realistic. 
Thus, the methods that only make use of global spatial features may not work well on the models: 1) that synthesize entire images with refined global qualities and subtle artifacts in local regions; 2) that partially manipulate the image like swapping the face region only in recent DeepFakes. 

Inspired by the previous works~\cite{zhao2021multi, zhang2021multi}, we propose a two-branch model to combine global and local features to detect diverse types of AI synthesized images. The global branch focuses on extracting overall structural features from the whole image. The local branch extracts fine-grained artifacts from multiple patches. Instead of using random or manually pre-defined patches, we design a Patch Selection Module~(PSM) to automatically select informative patches. The features from two branches are fused by an Attention-based Feature Fusion Module~(AFFM) and fed into a binary classifier. 


    
    

Our main contributions can be summarized as follows:
\vspace{-0.3cm}
\begin{itemize}
    \item We propose a two-branch framework to combine global features from the whole image with the local features from informative patches to enhance the generalization ability of AI-synthesized image detection. 
    \vspace{-0.25cm}
    \item We design an attention-based patch selection module to automatically select multiple patches for local subtle artifacts extraction without any manual annotations. This module facilities the model to locate locally manipulated regions more efficiently. \vspace{-0.25cm}
    \item We collect a large-scale and highly diverse dataset with different resolutions, various manipulation types, and extensive synthesis models to evaluate our method. Experimental results demonstrate the outstanding generalization ability of our method. 
\end{itemize}

\begin{figure}[!t] 
\centering 
\includegraphics[width=0.48\textwidth]{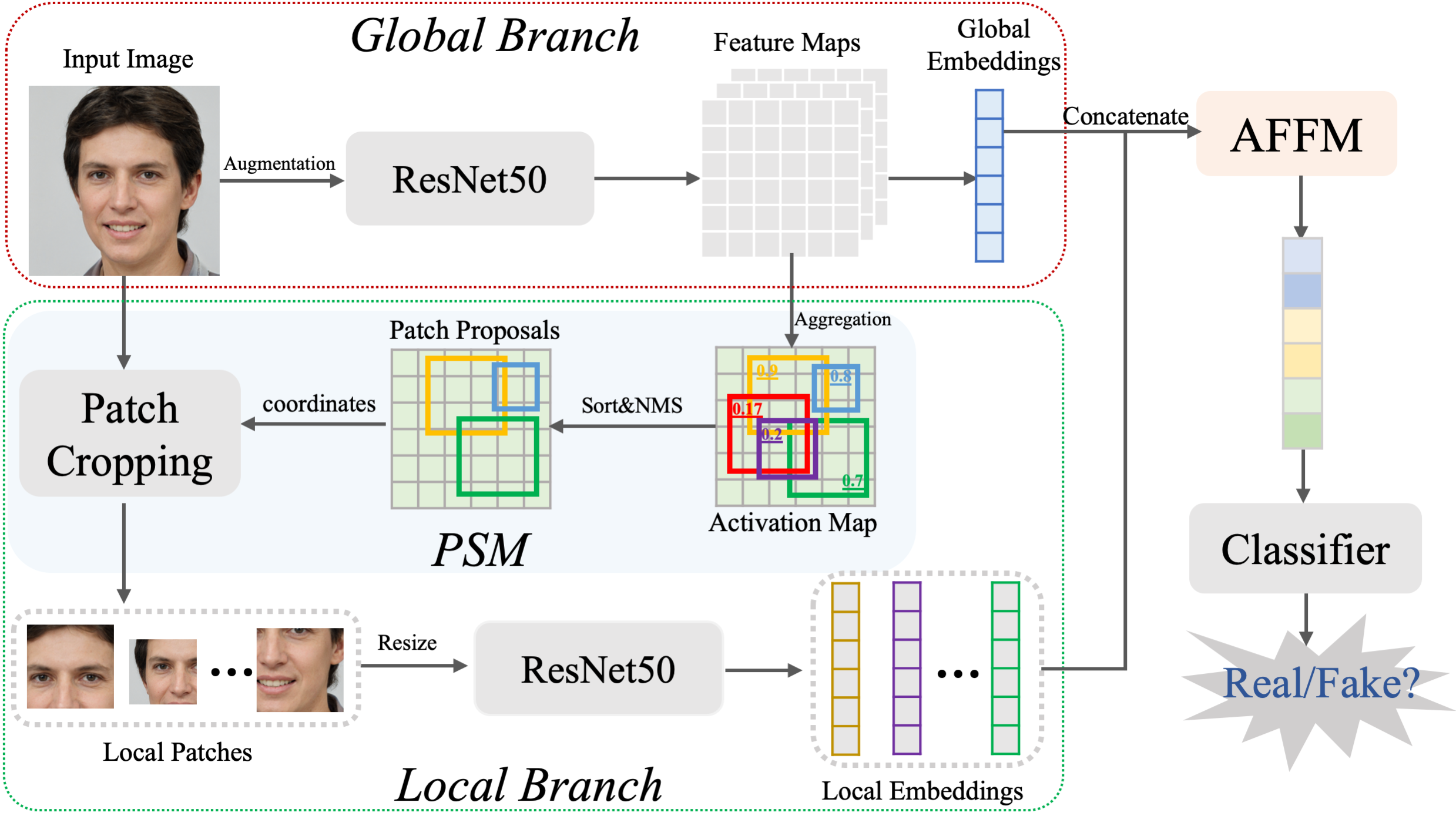}
\vspace{-0.8cm}
\caption{\small \em \textbf{The architecture of our model.} Global branch extracts spatial feature and local branch extracts subtle feature from local patches selected by Patch Selection Module~(PSM). The Attention-based Feature Fusion Module~(AFFM) fuses global and local features for the classification.} 
\label{fig:mainfig} 
\vspace{-0.7cm}
\end{figure}

\vspace{-0.75cm}
\section{Related Work}
\label{sec:relatedwork}
\vspace{-0.35cm}

\noindent \textbf{Fake Image Generation}.
The generative adversarial networks~(GANs) are usually used to synthesize entire images. Unconditional GANs take a random vector as input and no class labels are needed for generative modeling. Examples of widely used unconditional GAN models include StyleGAN-V1/2/3~\cite{karras2019style,karras2020analyzing,karras2021alias}, which can generate realistic high-resolution images. Conditional GANs models such as CycleGAN~\cite{zhu2017unpaired} often take an image as input and generate the specific class of image with different object, style, etc. For partially manipulated images/videos such as face swaps, puppetry and attributes manipulation, most methods are based on the auto-encoder models~\cite{mirsky2021creation}. Several large scale datasets such as FaceForensic++~\cite{rossler2019faceforensics++} and Celeb-DF \cite{li2020celeb} are proposed.

\noindent \textbf{Fake Image Detection}. There have been many active developments in methods to distinguish between real and synthetic/manipulated images. A common approach is to train a binary classification detector on whole image with deep neural network models~\cite{wang2020cnn,gragnaniello2021gan,asnani2021reverse,zhao2021multi,guo2021robust}. These methods focus on global feature extraction from the whole image and are often not sensitive to the subtle local artifacts. To obtain more generalized features for diverse fake manipulation, recent studies have emphasized the significance of informative patches~(local features) for detection and localization~\cite{chai2020makes,schwarcz2021finding,zhang2021thinking}. Specifically, the study in \cite{chai2020makes} proposes a patch-based classifier with limited receptive fields and obtains excellent generalization on the entire face forgery datasets. The method in \cite{schwarcz2021finding} also focuses on various facial parts and evaluate the performance on each individual part. However, this method puts too much emphasis on local artifacts and ignores global ones, which may not generalize well to both entire forgery and partially forgery detection. 

\vspace{-0.55cm}
\section{Method}
\label{sec:method}
\vspace{-0.25cm}
\vspace{-0.1cm}
\subsection{Overall Architecture}
\vspace{-0.2cm}

\begin{table*}
\renewcommand\arraystretch{1.15}
\center
\caption{\label{tb:datasets} \small \em Details of our highly-diverse testing dataset.} 
\scalebox{0.72}{
\begin{tabular}{c  c  c  c  c  c}
 \hline
 \textbf{Family} & \textbf{Models} &  \textbf{\# Images (Real/Fake)} & \textbf{Object} & \textbf{ Resolution} \\
 \hline
 \textbf{Contiditonal GANs} & \begin{tabular}{@{}c@{}}GauGAN\cite{park2019semantic}, StarGAN\cite{choi2018stargan}, \\ CycleGAN\cite{zhu2017unpaired}, Pix2Pix\cite{isola2017image}\end{tabular} & 19,343 (10,023/9,320) & \begin{tabular}{@{}c@{}} Face \& Non-face \\ (bedroom,apple,winter,etc.)\end{tabular} & $256\times256$\\ 
 \hline
 \textbf{Uncontiditonal GANs} & \begin{tabular}{@{}c@{}}ProGAN\cite{karras2017progressive}, StyleGAN\cite{karras2019style}, \\ StyleGAN2\cite{karras2020analyzing},  StyleGAN3\cite{karras2021alias}, \\ BigGAN\cite{brock2018large}, WGAN\cite{arjovsky2017wasserstein}, VAEGAN\cite{larsen2016autoencoding}\end{tabular}  & 71,750 (27,335/44,415)  & \begin{tabular}{@{}c@{}} Face \& Non-face \\ (car,church,horse,etc.)\end{tabular}   &  $160\times160 \sim$  $5k\times8k$\\
 \hline
 \textbf{Perceptual Loss} & CRN\cite{chen2017photographic}, IMLE\cite{li2019diverse} & 25,528 (12,764/12,764) & Non-face(street) & $512\times256$\\
 \hline
 \textbf{Low-level Vision} & SITD\cite{chen2018learning}, SAN\cite{dai2019second} & 798 (399/399) & Face\&Non-face(flower, animal, etc.) & $480\times320 \sim 6k\times4k$\\
 \hline
 \textbf{DeepFakes} & FF++\cite{rossler2019faceforensics++}, Celeb-DF\cite{li2020celeb} & 7,097 (3,531/3,566) & Face & $256\times256 \sim 944\times500$\\
 \hline
 \textbf{Others} &  Transformers\cite{esser2021taming}, BGM\cite{MODNet} & 3,908 (1,438/2,470) & Face & $256\times256 \sim 1024\times1024$ \\
 \hline
\end{tabular}}
\vspace{-0.55cm}
\end{table*}
The overall structure of our two-branch model is illustrated in Fig.~\ref{fig:mainfig}. 
Taking an image as input, the global branch generates a stack of global feature maps and a global feature embedding using the backbone network (e.g, ResNet50~\cite{he2016deep}). The Patch Selection Module~(PSM) predicts the coordinates of most informative patches that are then cropped from the input image and fed into local branch to obtain the local features. An Attention-based Feature Fusion Module~(AFFM) is further designed to fuse the global and patch embeddings.

\vspace{-0.45cm}
\subsection{Global Branch}
\vspace{-0.2cm}
The global branch in the model is designed for high-level global spatial feature learning. It takes an image as input and employs the CNN feature extractor to obtain the global features maps from last convolutional layer, denoted as $F \in \mathbb{R}^{C \times H \times W}$ with $C$ channels and $ H \times W$ spatial size. The global embeddings are acquired by inputting $F$ into average pooling and fully connected layers. The global features contain the spatial and structural information from the input image.

\vspace{-0.45cm}
\subsection{Local Branch with Patch Selection Module}
\vspace{-0.2cm}
Our local branch is designed to locate the most informative patches, from which the discriminative local features are extracted. Assuming that the regions with higher energy in global feature maps usually are more informative for classification task, we design a Patch Selection Module~(PSM) module to locate more useful patches from global feature maps. The PSM module extracts the coordinates of the most informative patches in the input image by calculating the patch scores from the global feature maps $F$. 

To calculate the patch scores, we first estimate the activation map $A$ by aggregating the $F$ in the channel dimension as $A(x,y) = \sum_{j=1}^{C} F^j(x,y)$.
Then we calculate the average score in a window on the activate map to indicate the ``informativeness'' of the corresponding patch in the input image. Specifically, we set the windows with $H_{w} \times W_{w}$ spatial size to slide on the activate map $A$ and calculate the score of each window $S_{w}$ as following:

\vspace{-0.3cm}
\begin{equation}
\label{eqn:sw}
S_{w} = \frac{\sum_{x=0}^{W_{w}-1}\sum_{y=0}^{H_{w}-1}A(x, y)}{W_{w} \times H_{w}}
\vspace{-0.2cm}
\end{equation}

By sorting the scores, a fixed number of windows with higher scores are selected as our patch proposals after using non-maximum suppression~(NMS) to reduce patch redundancy. Coordinates of these patch proposals are then mapped back to the original input image. After patch cropping, the most informative local patches are selected, which are fed into the CNN to extract local embeddings.

\vspace{-0.45cm}
\subsection{Attention-based Feature Fusion Module}
\vspace{-0.2cm}
To efficiently combine the global and local features for final classification, an Attention-based Feature Fusion Module~(AFFM) based on multi-head attention mechanism~\cite{vaswani2017attention} is designed. After concatenating the global embeddings from global branch and several patch embedddings from local branch, the AFFM learns correlations among these embeddings and combine them as a whole feature with the learned weights for further classification. Specifically, the concatenated embeddings are inputted into AFFM as the query $\mathrm{Q}$, the key $ \mathrm{K}$, and the value $\mathrm{V}$. Multi-head attention with $n$ heads is used. The $i$th attention head $\mathrm{h}_i (i\in [1,n])$ is computed with the scaled dot product attention as $\mathrm{h}_{i}=softmax(\frac{{\mathrm{W}_i}^Q\mathrm{Q}\times{{(\mathrm{W}_i}^K\mathrm{K})}^T}{\sqrt{d}})\times {\mathrm{W}_i}^V\mathrm{V}$, where ${\mathrm{W}_i}^Q$, ${\mathrm{W}_i}^K$, ${\mathrm{W}_i}^v$ are parameter matrices and $d$ is the dimension of $K$. The attention scores of all heads are concatenated as the input of a linear transformation to obtain the value of multi-head attention as $MHA(Q,K,V) = Concat(\mathrm{h}_i)\mathrm{W}^o (i \in [1,n])$, where $\mathrm{W}^o$ is also a parameter matrix. The fused one dimensional vector from $MHA(Q,K,V)$ is taken as the learned feature and inputted into classifier for the final prediction.



\begin{table*}
\renewcommand\arraystretch{1.15}
\center
\footnotesize
\caption{\label{tb:map} \small \em \textbf{Evaluation results on the Seen and Unseen models.}  We show the mAP of each family, Total mAP of all the models and Global AP. The highest value is highlighted in black. Baselines are listed in the first three rows followed by two ablations~(Our\_RdmCrop and Our\_Rsz448). The proposed method Our\_PSM is listed in the last row. } 
\scalebox{0.97}{
\begin{tabular}{ c | c | c  c  c c c c |c |c}
 \hline
   \multirow{2.5}*{\diagbox{Method}{Family}} & Seen & \multicolumn{6}{c|}{Unseen} & Total & Global\\
  \cline{2-8}
        & ProGAN & \thead{Conditional\\ GANs} & \thead{Unconditional \\GANs} & \thead{Perceptual\\ Loss} & \thead{Low-level\\ Vision} & DeepFakes & Others & mAP & AP\\
 \hline
  CNN\_aug\cite{wang2020cnn} & \textbf{100} & 94.782 & 88.100 & 98.374 & 82.027 & 69.576 & 71.846 & 86.389 & 93.578 \\
 \hdashline
  No\_down\cite{gragnaniello2021gan} & \textbf{100} & 91.294 & 90.438 & 93.820 & \textbf{89.853} & \textbf{72.033} & 81.445 & 88.063 &  91.593\\
  \hdashline
  Patch\_forensics\cite{chai2020makes} & 94.620 & 89.245 & 69.286 & 99.472 & 69.661 & 57.424 & 57.902 & 73.979 &  70.149\\
 \Xhline{2\arrayrulewidth}
  Our\_RdmCrop & \textbf{100} & 96.039 & 88.775 & 98.719 & 71.773 & 63.154 & 82.357 & 86.825 & 94.457 \\
 \hdashline
  Our\_Rsz448 & \textbf{100} & 97.286 & 92.712 & \textbf{99.930} & 74.340 & 68.960 & 80.143 &  88.813 & 95.717 \\
 \hdashline
 Our\_PSM & \textbf{100} & \textbf{98.305} & \textbf{95.485} & 99.147 & 85.104 & 70.109 & \textbf{87.493} & \textbf{91.732} & \textbf{96.906} \\
 \hline
\end{tabular}}
\vspace{-0.55cm}
\end{table*}

\vspace{-0.35cm}
\section{Experiments}
\vspace{-0.15cm}

\vspace{-0.15cm}
\subsection{Datasets}
\vspace{-0.15cm}
The training set provided in~\cite{wang2020cnn} was used for our model training. It comprises 362K real images from $20$ object classes of the LSUN dataset and 362K images generated by ProGAN~\cite{karras2017progressive}. The spatial size of all training images is $256 \times 256 $. For the testing dataset, we composed a dataset of synthetic images generated with 19 various generation models based on several existing datasets~\cite{wang2020cnn, asnani2021reverse,li2020celeb}, considering both whole image synthesis (GANs) and partial manipulations (face swaps). We also included several recently released models such as BGM~\cite{MODNet}, StyleGAN3~\cite{karras2021alias}, Transformers~\cite{esser2021taming}, Celeb-DF~\cite{li2020celeb}, etc. 
These $19$ models were further divided into $6$ families based on the model structures (see more details in Table~\ref{tb:datasets}). A total of $128,424$ images have been collected, with a high diversity in generation model, data object and image resolution. To the best of our knowledge, this is the most comprehensive and diverse dataset for AI synthesized image detection evaluation.

\vspace{-0.35cm}
\subsection{Implementation Details}
\vspace{-0.15cm}

Our model is implemented with Pytorch~\cite{paszke2019pytorch} and trained with $64$ batch size and $0.0001$ base learning rate. The Adam optimizer is used with the loss function of BCE. ResNet-50~\cite{he2016deep} pre-trained with ImageNet is utilized as backbone to extract global and local features. Input image is resized to $224 \times 224$ for our global feature extraction branch. Before resizing, $10\%$ training samples are selected randomly in each training epoch and augmented with Gaussian blur with Sigma parameter $\sigma$ $\sim$ Uniform $[0, 3]$ and JPEG-ed with quality $\sim$ Uniform $\{30, 31, . . . , 100\}$. Feature maps from layer $Conv_5b$ in ResNet-50 are inputted into PSM to guide informative patches selection. In PSM implementation, two sliding windows $H_w \times W_w = 3 \times 3$ and $H_w \times W_w = 2 \times 2$ are used to capture artifacts from different resolution on activation map, and the sliding step is set as $1$. $3$ patch proposals from window size $3 \times 3$ are selected and mapped into $224 \times 224$ patches on the input image. Similarly, $3$ patch proposals from window sizes $2 \times 2$ are mapped into $112 \times 112$ patches. These $6$ patches are resized into $224$ and then fed into ResNet-50 for local embeddings extraction. The dimension of extracted global embeddings and patch embeddings is $128$. In AFFM, $3$ attention layers with  $4$ heads are utilized to fuse all the extracted features. 

As for baselines, we use CNN\_aug~\cite{wang2020cnn}, No\_down~\cite{gragnaniello2021gan} and Patch\_forensics~\cite{chai2020makes}. We follow the settings in their paper and train them on our training dataset from scratch. Note that for Patch\_forensics, we use their released weights. For metric, we utilize average precision (AP) to evaluate our model and the baselines following recent works~\cite{wang2020cnn}. We report the mean AP~(mAP) by averaging the AP scores over each family. 

\vspace{-0.4cm}
\subsection{Evaluation for AI-synthesized Image Detection}
\vspace{-0.2cm}

Table~\ref{tb:map} reports the mAP of each family of the baselines and the proposed Our\_PSM. The total mAP among all the testing models are also listed. More importantly, to accurately measure the model performances in a real-world scenario where the source of the images is unknown, we mixed up all the images in the testing dataset and reported the global AP on the mixed dataset. All the methods achieve $100\%$ accuracy on the seen ProGAN model except Patch\_forensics, which is limited to face object detection and not works well on our testing data with diverse objects. For the unseen models, compared with three baselines, Our-PSM achieves the best mAP values among most families and performs slightly worse on Low-level Vision and DeepFakes than the No\_down method. We conjecture that the resizing operations of our model on the high-resolution images in Low-level Vision and DeepFakes families will make the artifacts more difficult to detect. 
Overall, Our\_PSM achieves the best performance in terms of Total mAP and Global AP. This demonstrates the better generalization ability of our model than the methods that only use whole image information or only focus on features from small patches.


To further evaluate the effectiveness of the proposed Patch Selection Module~(PSM), we also show the mAP obtained by replacing our PSM with two popular patch selection methods in the bottom $3$ lines of Table~\ref{tb:map}. In Our\_RdmCrop, the global feature extraction and AFFM stay the same except that the PSM is replaced with random patch cropping on the input image. Similarly, in Our\_Rsz448, we replace the PSM with resizing the input image to $448$ first and cropping five $224 \times 224$ patches from four corners and center. As shown in Table~\ref{tb:map}, Our\_PSM outperforms other two patch selection methods in all families except in Perceptual Loss. However, considering the overall performance, Our\_PSM demonstrates the best performance because our patch selection focuses on the most distinguishable patches and achieves the balance between using all the information of image and training efficiency. 
\begin{figure}[htb]

\begin{minipage}[b]{0.47\linewidth}
\vspace{-0.2cm}
  \centering
  \centerline{\includegraphics[width=3.2cm]{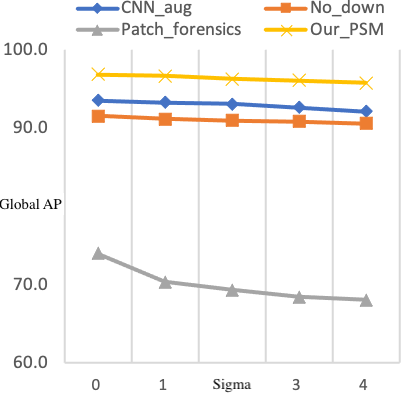}}
  \centerline{\small \em (a) Robustness to Blurring}\medskip
\end{minipage}
\begin{minipage}[b]{.47\linewidth}
\vspace{-0.2cm}
  \centering
  \centerline{\includegraphics[width=3.2cm]{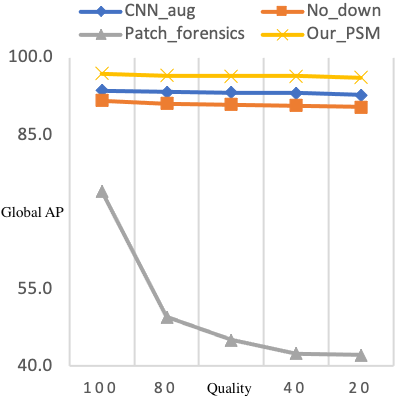}}
  \centerline{\small \em (b) Robustness to JPEG}\medskip
\end{minipage}
\hfill
\vspace{-0.6cm}
\caption{\small \em \textbf{Comparison of detection robustness to post-processing.}}
\label{fig:robustness}
\vspace{-0.4cm}
\end{figure}

We also evaluate the robustness of our method and the baselines on the images post-processed with Gaussian Blur and JPEG Compression. Fig.~\ref{fig:robustness} shows that our model achieves good robustness to these post-processing operations even on such a diverse unseen dataset. This indicates that data augmentation during training process may help to improve the robustness.





\vspace{-0.75cm}
\section{Conclusion}
\vspace{-0.45cm}
\label{sec:conclusion}

In this work, we propose a two-branch model that fuses global and informative local features for generalized AI-synthesized image detection. Global branch learns the whole image structure information, and the local branch extracts local subtle features from the informative regions selected by the proposed PSM. Multi-head attention is finally used to fuse these complementary features for a binary classification. Our model learns to focus on the most discriminative local patches with the help of the global high-level features in an unsupervised way, which allows us to fuse the local and global information more efficiently. Extensive experiments on the highly diverse testing dataset that contains $19$ models demonstrate that our proposed method achieves high accuracy and good generalization ability. In future work, we will investigate a differentiable patch selection module so that the patches can be learned more efficiently. 

Acknowledgements: This work is supported by the US Defense Advanced Research Projects Agency (DARPA) Semantic Forensic (SemaFor) program, under Contract No. HR001120C0123. We thank SemaFor for providing datasets for our training and testing. 



\footnotesize
\bibliographystyle{IEEEbib}
\bibliography{refs}

\end{document}